\documentclass[runningheads]{llncs}
\usepackage[english]{babel}
\usepackage{graphicx}
\usepackage[table]{xcolor}
\usepackage{multirow}
\usepackage{hyperref}
\usepackage{cite}

\addto\extrasenglish{%
}

\begin{document}
\title{
Analyzing An After-Sales Service Process Using Object-Centric Process Mining: A Case Study
}
\titlerunning{Object-Centric Process Mining: Case Study}
%
\author{
Gyunam Park\inst{1}
\and
Sevde Aydin\inst{2}
\and
C\"{u}neyt U\u{g}ur\inst{3}
\and
Wil M. P. van der Aalst\inst{1}
}
\authorrunning{G. Park et al.}
%
\institute{Process and Data Science Group (PADS), RWTH Aachen University \\ \email{\{gnpark,wvdaalst\}@pads.rwth-aachen.de} \and Gebze Technical University \\ \email{s.aydin2019@gtu.edu.tr} \and AI and Process Automation Department, Borusan Cat - R\&D, Digital and Technology \\ \email{cugur@borusan.com}}
\maketitle              

\begin{abstract}
Process mining, a technique turning event data into business process insights, has traditionally operated on the assumption that each event corresponds to a singular case or object. However, many real-world processes are intertwined with multiple objects, making them object-centric. This paper focuses on the emerging domain of object-centric process mining, highlighting its potential yet underexplored benefits in actual operational scenarios. Through an in-depth case study of \textit{Borusan Cat}'s after-sales service process, this study emphasizes the capability of object-centric process mining to capture entangled business process details. Utilizing an event log of approximately 65,000 events, our analysis underscores the importance of embracing this paradigm for richer business insights and enhanced operational improvements.
\end{abstract}

\section{Introduction}
Process mining leverages event data from operational processes to gain insights~\cite{Aalst16}.
This includes techniques such as process discovery, which automatically derives process models from event data; conformance checking, which compares the recorded event log with the process model; process enhancement, which augments the process model with frequency and performance details; and predictive process monitoring, which foresees the remaining time and potential risk of an ongoing case.
Companies like Siemens, Uber, BMW, and Bosch have effectively employed process mining, achieving substantial savings amounting to millions of Euros~\cite{reinkemeyer20}.

In process mining, there is a prevalent assumption that each event corresponds directly to a singular, specific case.
Consider a healthcare scenario: an event, like registration, refers directly to a single patient.
Yet, this assumption often does not hold in real-world contexts.
Instead, many business processes, in reality, are shaped by the interactions of several intertwined objects, making them \textit{object-centric}~\cite{math11122691}.
For example, an omnipresent Purchase-To-Pay (P2P) process encompasses multiple object types, including purchase orders, goods receipts, and invoices.
An event within such processes can relate to multiple objects of distinct types.
For instance, event \textit{verify goods receipts} in a P2P process might correspond to multiple goods receipts and the associated purchase order.
Similarly, an event \textit{three-way matching} relates to an invoice, its relevant goods receipts, and the corresponding purchase order to confirm the invoice's amount.

Object-centric process mining deviates from the conventional assumption that each event is tied to only one case or object.
Rather, it allows an event to connect with multiple objects.
This approach grants analysts the adaptability to select their preferred object and event perspectives for various analyses.
It also captures the complexity of object interactions, enabling a deeper analysis of intertwined business processes, and leverages multi-dimensional data models for a more precise representation of business processes~\cite{OCPM-white-paper}.

In recent years, various tools and techniques tailored for object-centric process mining have emerged~\cite{DBLP:journals/is/GhilardiGMR22,DBLP:books/sp/22/Fahland22,DBLP:conf/bpm/RebmannRA22}.
However, while this paradigm holds great potential, the corresponding tools and techniques have not been widely applied.
Therefore, there is a need for documented real-life applications of these concepts.

This paper presents a case study using the after-sales service process of \textit{Borusan Cat} within \textit{Borusan Group}.
As one of \textit{Caterpillar Inc.}'s 160 dealers - a global front-runner in construction and mining equipment manufacture - \textit{Borusan Cat} operates in six countries with a workforce exceeding 3,000.
They cater to three primary sectors: construction, mining, and energy \& transportation within these nations.
Our study utilizes an event log containing around 65,000 events, aiming to present the findings, underscoring the potential of object-centric process mining in a service process context.

The structure of this paper is as follows:
\autoref{sec:context} details the background of the case study.
Next, \autoref{sec:planning} outlines the planning of the case study, introducing the process and defining the goals for analysis.
Afterward, \autoref{sec:dataExtraction} describes the extraction of the object-centric event log.
Then, \autoref{sec:preProcessing} introduces our data preprocessing strategies.
In \autoref{sec:processMiningAnalysis}, we delve into the mining and analysis phase.
Next, \autoref{sec:improvement} discusses both implemented and prospective improvement strategies.
Finally, \autoref{sec:conclusion} concludes the paper.

\section{Background}
\label{sec:context}

This section presents the context of the company and the process where the proposed case study is performed.
Moreover, we introduce a process mining methodology used to conduct our case study.

\subsection{Company}
\label{subsec:company}
The Borusan Group\footnote{\url{https://www.borusan.com/en/home}} stands as one of T\"{u}rkiye’s most distinguished conglomerates, boasting a workforce of more than 12,000 in 12 countries spanning three continents.
As a dealer of \textit{Caterpillar Inc.} and numerous leading brands, \textit{Borusan Cat} serves the construction, resource, energy, and transportation industries.
With a team of more than 3,000 employees in six countries, the company provides machinery, generators, and spare parts sales, complemented by after-sales support.
Beyond offering Cat construction equipment and generator rental services, \textit{Borusan Cat} delivers holistic solutions to enterprises in the construction, mining, energy, marine, and oil sectors.

Driven by its goals of amplifying operational excellence and enhancing customer satisfaction, \textit{Borusan Cat}'s process excellence team has analyzed the after-sales service process, focusing on Key Performance Indicators (KPIs) such as chargeable working hours, deviations between actual and planned hours, technician turnaround durations, technician productivity, their time optimization, and operational effectiveness at client locations based on daily technician activities.
These established KPIs serve as a compass to oversee the after-sales service process.

\subsection{Process Description}
\label{subsec:descriptionProcess}

\begin{figure}[!ht]
\centering
\includegraphics[width=0.6\textwidth]{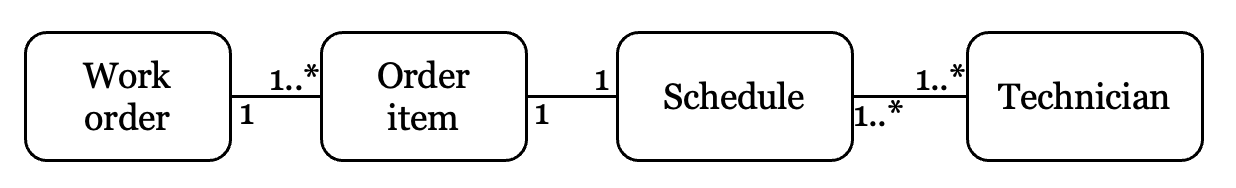}
\caption{A data model of the after-sales service process}
\label{fig:data_model}
\end{figure}

\autoref{fig:data_model} shows a data model of the after-sales service process.
A customer issues a work order that may encompass multiple order items, e.g., each item for an issue.
A schedule is created for each order item and can be allocated to several technicians, while a technician is involved in multiple schedules.
In other words, the relationship between schedules and technicians is many-to-many.

\begin{figure}[!ht]
\centering
\includegraphics[width=1\textwidth]{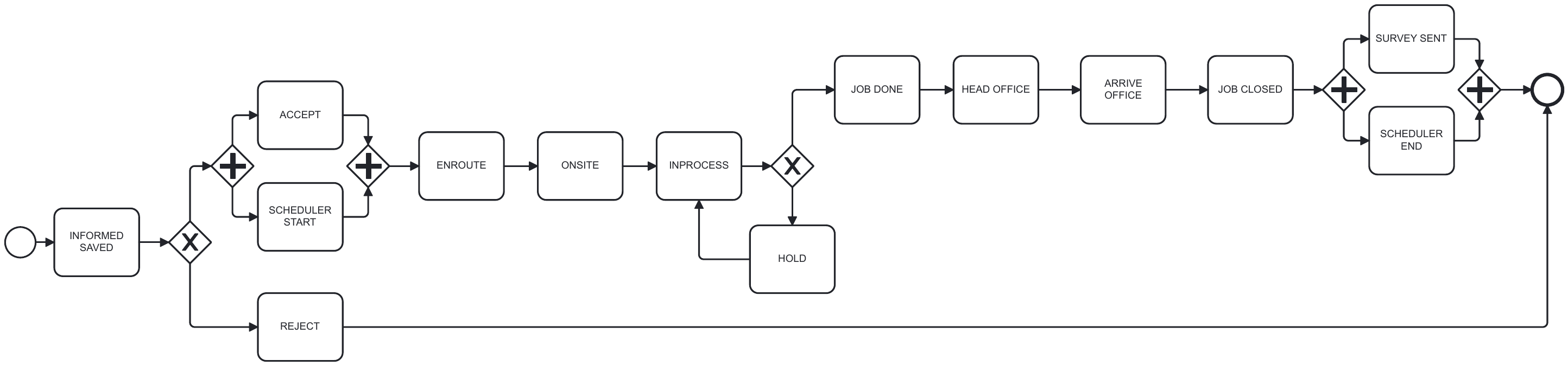}
\caption{Reference process model in BPMN notations}
\label{fig:bpmn}
\end{figure}

\autoref{fig:bpmn} shows the reference process model for the process where a single technician handles a single schedule.
Note that the reference model does not describe the process where multiple technicians handle multiple schedules.
First, a schedule is assigned to a technician based on customer information (name, location, and job description) and work order details (order number and item number) from the SAP system.
Upon the start time of the assigned schedule, the system initiates the SCHEDULER START activity.
When the end time of a technician's schedule approaches, the system activates the SCHEDULER END activity.
Technicians must complete all customer-related activities within this scheduled timeframe.
Work completed within this period is deemed \textit{normal work}, while any extra work outside this timeframe is termed \textit{overwork}.

Technicians can either accept or decline the assigned schedule for various reasons (e.g., adverse weather and equipment unavailability). 
An ACCEPT activity is recorded for acceptance, and REJECT for declination.
After accepting, the technician sets out for the customer's location, marking the ENROUTE status on their mobile app.
After arriving at the customer's location, the technician updates the status to ONSITE.
The technician begins to work on the order, indicating this with the INPROCESS status.
If a technician pauses their work, they select the HOLD status.
Upon finishing the job, the technician selects JOB DONE.
The technician updates the HEAD OFFICE status as they head back.
Upon return, the technician marks ARRIVE OFFICE.
If all tasks for a customer are finished on the same day, the technician updates the status to JOB CLOSED.
After service provision, technicians send an evaluation survey to customers, denoted by the SURVEY SENT activity.

\subsection{Methodology}
To apply process mining successfully, the process mining discipline provides several project methodologies aiming at supporting the application of process mining in organizational contexts. For instance, the L\textsuperscript{*} life-cycle model~\cite{Aalst16} and the Process Mining Project Methodology (PM\textsuperscript{2})~\cite{DBLP:conf/caise/EckLLA15} provide clear guidance to practitioners on how they implement process mining projects which aim to improve process performance and compliance to rules and regulations.

To apply object-centric process mining in an organizational setting, we extend ${PM}^{2}$ to guide the organization that seeks to apply object-centric process mining. The renewed methodology consists of five stages.

\begin{enumerate}
\item {\bf Planning}: this stage is to set up the project and determine the object-centric analysis' goals that need to be answered at the end of the project in a way that improves the process performance. 
\item {\bf Extraction}: this stage is to extract the event data from the information system and obtain the object-centric event log.
\item {\bf Data Preprocessing}: this stage is to prepare the event data so that the following mining and analysis techniques can produce optimal results.
\item {\bf Mining and Analysis}: this stage is to apply object-centric process mining techniques to the preprocessed event data and get insights into the interaction among various object types which answer the object-centric analysis' goals.
\item {\bf Improvement}: this stage transforms actionable insights into actual management actions that support the process to improve performance and compliance.
\end{enumerate}

In the following sections, we apply each step to the after-sales service process of \textit{Borusan Cat}.

\section{Planning}
\label{sec:planning}

First, the design of our case study was influenced by the insights given by our partners at the company.
We initiated a set of focused panel discussions to familiarize ourselves with the complexity of the scheduling process and how object-centric process mining could best be applied to it.

From our early interactions with company stakeholders, we discovered that they already employed a traditional process mining infrastructure.
This system was primarily centered on analyzing a single schedule by a single technician.
However, our discussions spotlighted the challenges and inadequacies of using traditional process mining approaches for analyzing the multi-faceted nature of the after-sales service process, i.e., the many-to-many relationships between schedules and technicians (cf. \autoref{fig:data_model}).
Moreover, the existing approaches did not provide insights into severe deviations and potential bottlenecks arising from the multi-faceted relationships between schedules and technicians.

To handle the gap, we have organized our objectives into three main pillars:
\begin{itemize}
    \item Transparency: To establish clear insight into the process, especially focusing on the multi-faceted interactions between schedules and technicians.
    \item Compliance monitoring: To detect and understand deviations from established business rules, focusing on the rules defined over the interaction of schedules and technicians.
    \item Efficiency analysis: To analyze potential bottlenecks in the process, focusing on the bottlenecks that occur at the interaction of schedules and technicians.
\end{itemize}

\section{Extraction}\label{sec:dataExtraction}
\begin{figure}[!ht]
\centering
\includegraphics[width=0.8\textwidth]{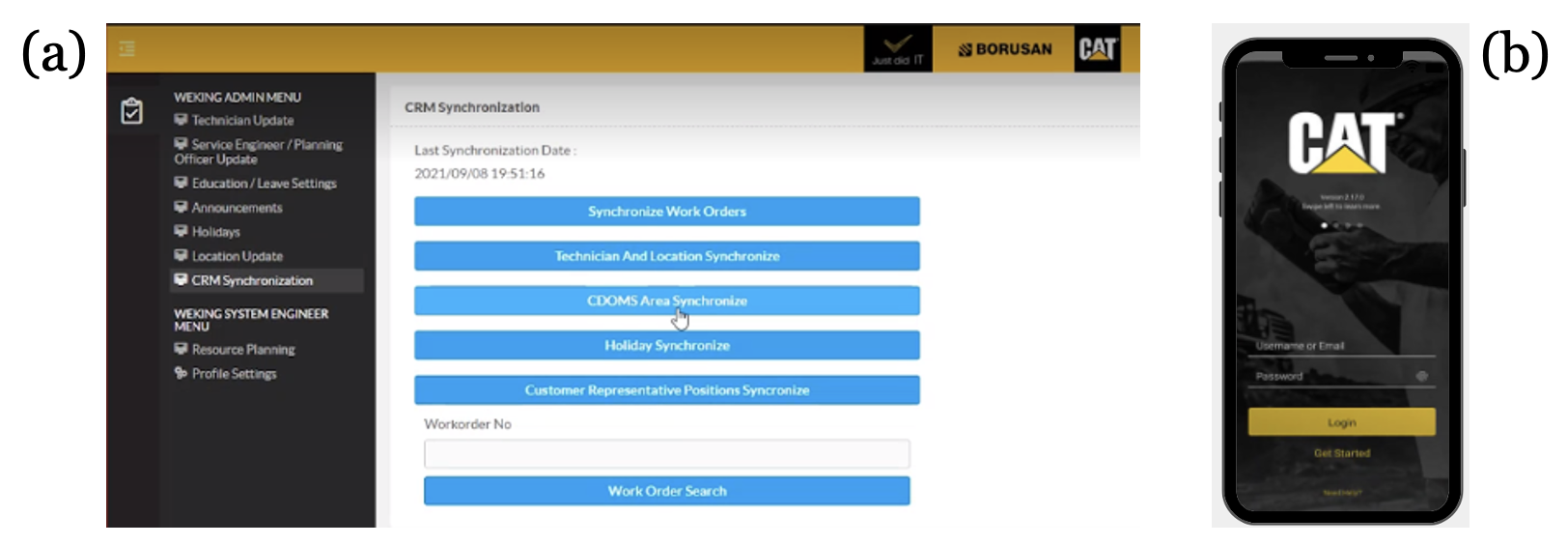}
\caption{\textit{WeKing} application that supports the after-sales service process: (a) web application and (b) mobile application}
\label{fig:system}
\end{figure}

We extracted an Object-Centric Event Log (OCEL) from a database tied to the application \textit{WeKing}.
The application serves as an integral tool for \textit{Borusan Cat}'s employees. 
Managers at \textit{Borusan Cat} use \textit{WeKing} web application (cf. \autoref{fig:system}(b)) to allocate schedules, whereas technicians use \textit{WeKing} mobile application (cf. \autoref{fig:system}(b)) for real-time status updates.
All these interactions are logged in an Oracle database.
From the database, we extract event data of schedules and technicians, adhering to the OCEL standard format\footnote{\url{http://www.ocel-standard.org/}}.

Our extraction concentrated on the events occurring in T\"{u}rkiye from 2023, specifically those related to external after-sales services where technicians visit customer sites. 
The details of an event, such as customer names and work order descriptions, were extracted. 

The resulting dataset comprises 65,774 events associated with 6,483 schedules and 173 technicians.
\autoref{tab:runningExampleTable} shows a fraction of the OCEL in tabular form.
The first entry, for instance, showcases an event labeled \emph{e1}, noting the activity \emph{ACCEPT} on \emph{2023-01-02 08:54}. This event refers to technician \emph{4006975} and schedule \emph{3948148}.

\begin{table}[!t]
\centering
\caption{Example object-centric event log of the after-sales service process represented as a table.}
\resizebox{0.5\textwidth}{!}{%
\begin{tabular}{|c|c|c|cc|}
\hline
{\bf Id} & {\bf Activity} & {\bf Timestamp} & {\bf Technician} & {\bf Schedule} \\
\hline
e1 & ACCEPT & 2023-01-02 08:54 & [4006975] & [3948148] \\
e2 & ENROUTE & 2023-01-02 08:54 & [4006975] & ~ \\
e3 & ONSITE & 2023-01-02 12:51 & [4006975] & ~ \\
e4 & INPROCESS & 2023-01-02 12:51 & [4006975] & [3948148] \\
... & ... & ... & ... & \\
\hline
\end{tabular}
}
\label{tab:runningExampleTable}
\end{table}

\section{Data Preprocessing}
\label{sec:preProcessing}
This section details the preprocessing steps employed on the extracted object-centric event log.
First, our dataset contains instances of unfinished process executions.
For instance, there are schedules that are still ongoing within the system.
To ensure the reliability and accuracy of our analyses, these incomplete executions are omitted from the event log.

Second, certain activities related to schedules and technicians adhere to a specific sequence.
For example, the activity ENROUTE should logically precede ONSITE.
This is because a technician would only arrive at a customer's site after initiating their journey. 
Similarly, the SCHEDULER START activity must come before SCHEDULER END. 
There were instances where these sequences were disrupted due to data recording anomalies.
The schedules and technicians involved in these irregularities were filtered out to maintain the integrity of our insights.

Third, our primary attention is on typical scenarios where a single technician is designated to a schedule.
We thus exclude schedules that involve multiple technicians from our analysis. 
Such occurrences are rare and often result from unintended actions by the technicians.
The resulting event log contains 57,601 events of 5,566 schedules and 169 technicians.

\section{Mining and Analysis}
\label{sec:processMiningAnalysis}
In this section, we conduct various analyses using the preprocessed object-centric event log.
First, we compute basic statistics to gain an initial understanding of the process and conduct single-viewpoint analyses focusing on schedules and technicians, respectively.
Next, we perform multi-viewpoint analysis using various object-centric process mining techniques to closely analyze the multi-faceted interaction between schedules and technicians.
Finally, based on the insights from the multi-viewpoint analysis, we conduct an in-depth analysis to analyze the implications and root causes of the insights to elicit improvements.

\subsection{Basic Statistics and Single Viewpoint Analysis}\label{subsec:single}
\begin{figure}[!ht]
\centering
\includegraphics[width=0.8\textwidth]{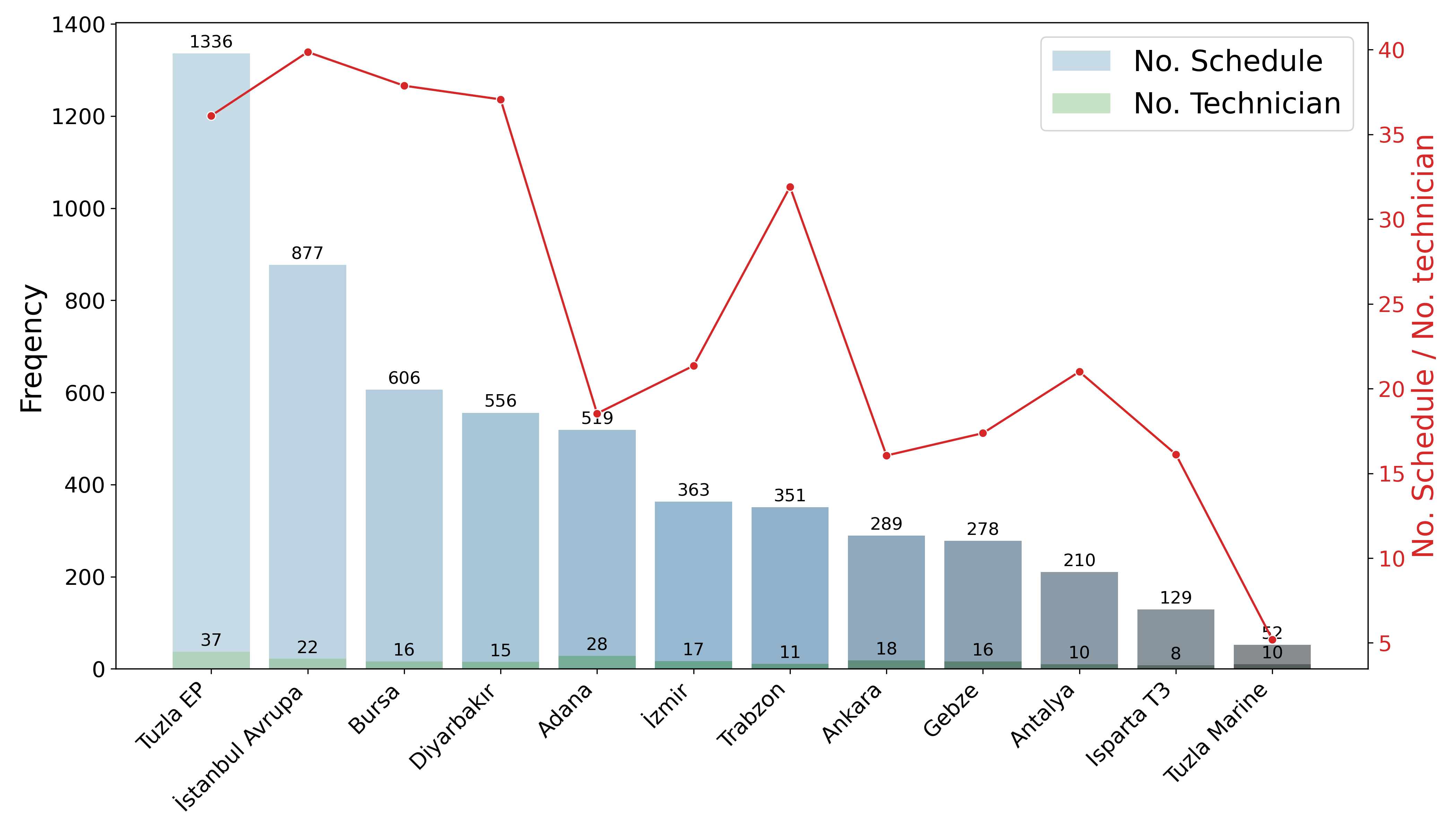}
\caption{Distribution of schedules and technicians across regions}
\label{fig:area_distribution}
\end{figure}

Initially, an explorative analysis was performed with the aim of obtaining a foundational understanding of the process.
\autoref{fig:area_distribution} shows the distribution of schedules and technicians across various regions.
Tuzla EP stands out with the largest number of schedules, i.e., 1,336, followed by Istanbul Avrupa, which has 877 schedules.
When considering the ratio of schedules per technician, Istanbul Avrupa leads with an average of 39.86 schedules assigned per technician.

\begin{figure}[!ht]
\centering
\includegraphics[width=1\textwidth]{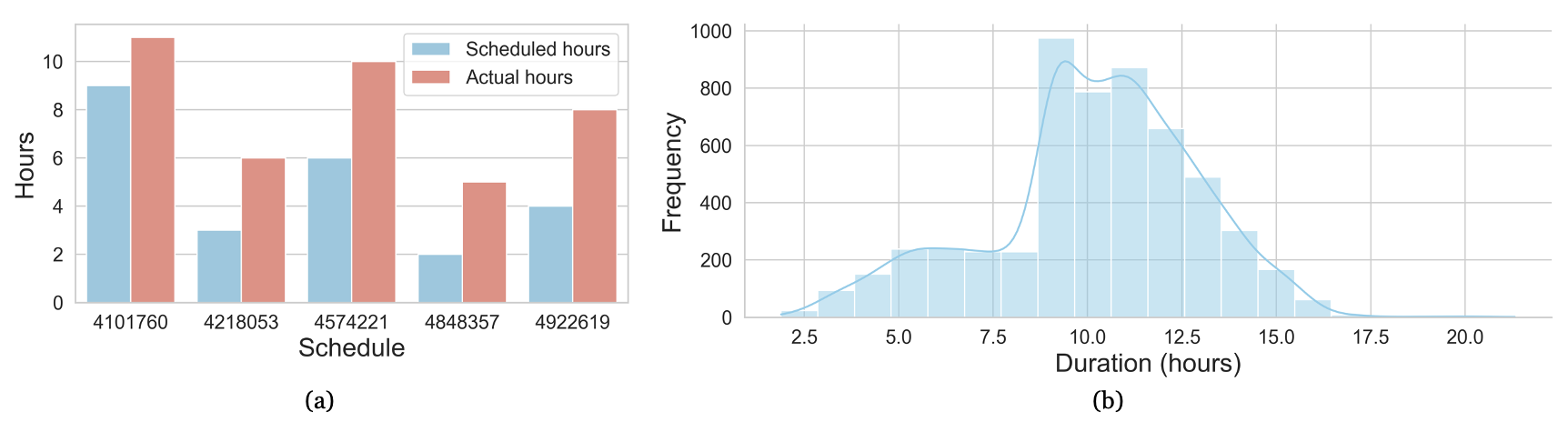}
\caption{(a) Overwork schedules and (b) Daily labor hours of a technician}
\label{fig:explorative}
\end{figure}

Next, we concentrated exclusively on the scheduling process.
In this process, \textit{scheduled hours} for every schedule can be computed as the time difference between the SCHEDULER START and SCHEDULER END activities.
Concurrently, \textit{actual hours} are inferred from the difference in time between ACCEPT and JOB CLOSED activities.
A schedule is considered as \textit{overwork} if the actual hours surpass the scheduled hours.
Out of 5,566 total schedules, 4,282 were identified as overwork.
For illustrative purposes, schedule 4101760, initially set for 9 hours, took a total of 11 hours, as demonstrated in \autoref{fig:explorative}(a).

Next, we transition to the technician's viewpoint.
First, we have observed that each technician has 340 activities on average.
It was deemed ineffective to use technicians as the case notion, given the extensive number of schedules each technician might possess.
To address this, we adopted a new case notion that embodies a technician's daily activities. 
In this context, each `case' characterizes a day's worth of a technician's events.

\autoref{fig:explorative}(b) unveils the daily labor hours of a technician.
Predominantly, their work stretches between 8 to 12 hours. 
However, in about 25\% of the instances, the work hours extend beyond 12.
Investigating deeper into this anomaly, we discover several underlying causes.
Among them, the technician's engagement in a HOLD activity emerged as a significant contributor.
Technicians engaged in such activities consistently showed a 3\% longer working duration compared to their counterparts without the HOLD activity.

\subsection{Multi-Viewpoints Analysis Using Object-Centric Process Mining}\label{subsec:multi}
In our preceding analysis, we examined individual viewpoints (perspectives) of the process related to schedules and daily technician tasks.
In the following analysis, we employ object-centric process discovery~\cite{DBLP:journals/fuin/AalstB20}, object-centric constraint monitoring~\cite{DBLP:conf/icpm/ParkA22}, and object-centric performance analysis~\cite{DBLP:conf/er/ParkAA22} to analyze the interplay between schedules and technicians.
For that, we use \textsc{ocpa}, a Python library supporting object-centric process mining~\cite{DBLP:journals/simpa/AdamsPA22}.

\begin{figure}[!ht]
\centering
\includegraphics[width=0.9\textwidth]{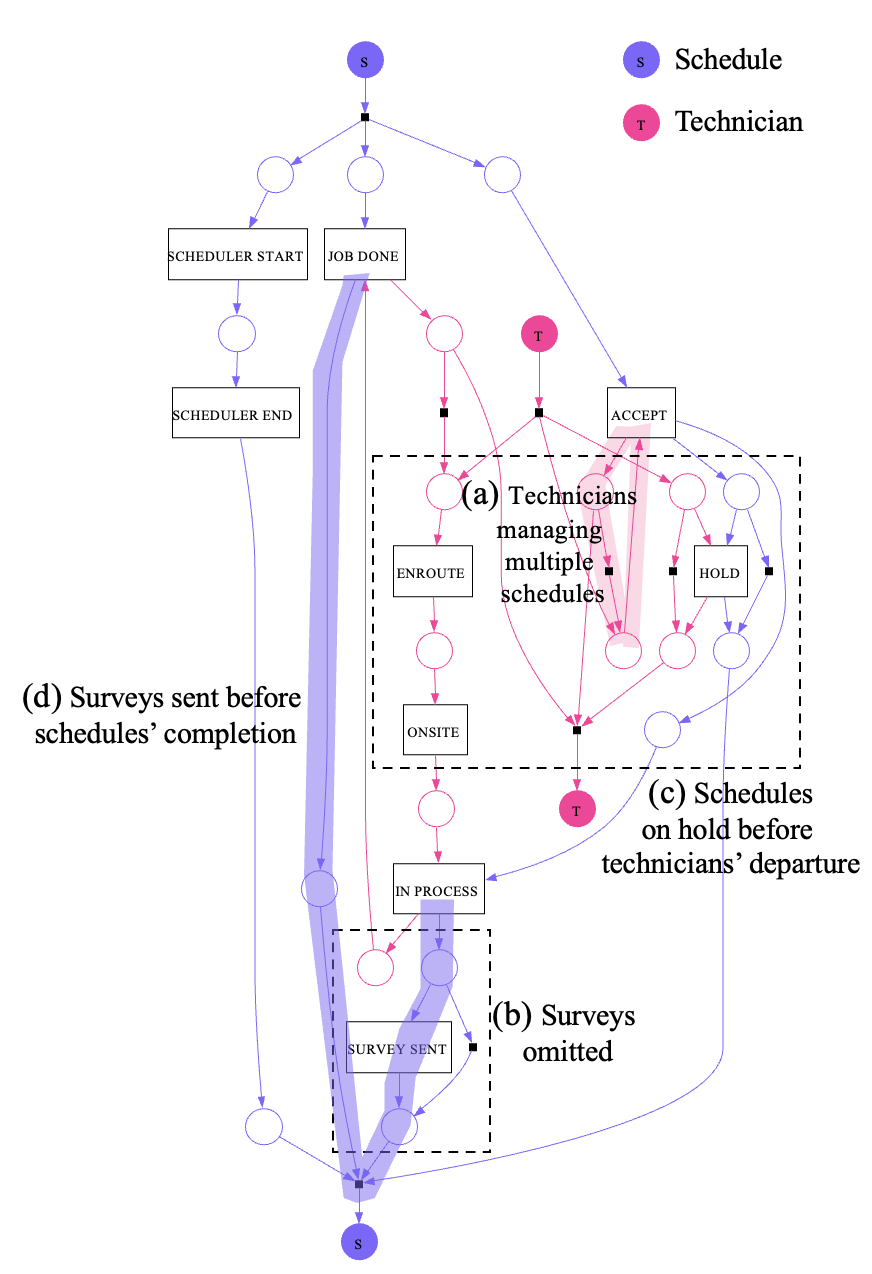}
\caption{Discovered process model as an object-centric Petri net~\cite{DBLP:journals/fuin/AalstB20}}
\label{fig:ocpn}
\end{figure}

First, we discover an object-centric Petri net from the extracted object-centric event log.
As shown in \autoref{fig:ocpn}, the discovered process model shows the interaction between the scheduling process (colored in purple) and the technician process (colored in pink).
Note that such interactions were not possible to observe in the reference model described in \autoref{fig:bpmn}.

Each schedule has a designated start and end time, represented as SCHEDULER START and SCHEDULER END, respectively.
Once accepted by a technician, the schedule is processed by the technician.
Upon completion of the schedule, a survey reaches the customer.

The technician's journey begins with accepting a schedule, followed by traveling to the designated site, and ultimately executing the task.
Given that a single technician might handle several schedules, this cycle could iterate beginning with the acceptance of a fresh schedule, as illustrated in \autoref{fig:ocpn}(a).
The cycle concludes when no additional schedules are in the pipeline.

Schedules and technicians intersect at multiple junctures.
Their interplay initiates when a technician accepts a schedule and terminates upon its completion.
In between, the technician works on the schedule and might occasionally pause it if needed.

The discovered process model unveils several deviations from the reference process model.
For instance, \autoref{fig:ocpn}(b) suggests that the customer survey is occasionally omitted. 
Additionally, as pinpointed in \autoref{fig:ocpn}(c), a schedule might be halted even before the technician reaches the site.
Moreover, \autoref{fig:ocpn}(d) indicates scenarios where the survey is dispatched prior to task completion.

Based on the insights derived from the process model, we have designed the following compliance rules and monitored their violations using the approach presented in~\cite{DBLP:conf/icpm/ParkA22}:
\begin{itemize}
\item Compliance rule 1: A survey should be sent per schedule.
\item Compliance rule 2: A schedule should not be halted until the technician arrives on-site.
\item Compliance rule 3: Surveys should be dispatched after the technician completes the scheduled task.
\end{itemize}
In 2023, compliance rule 1 was violated 2,222 times, while rules 2 and 3 were violated 144 and 124 times, respectively.

\begin{figure}[!ht]
\centering
\includegraphics[width=1\textwidth]{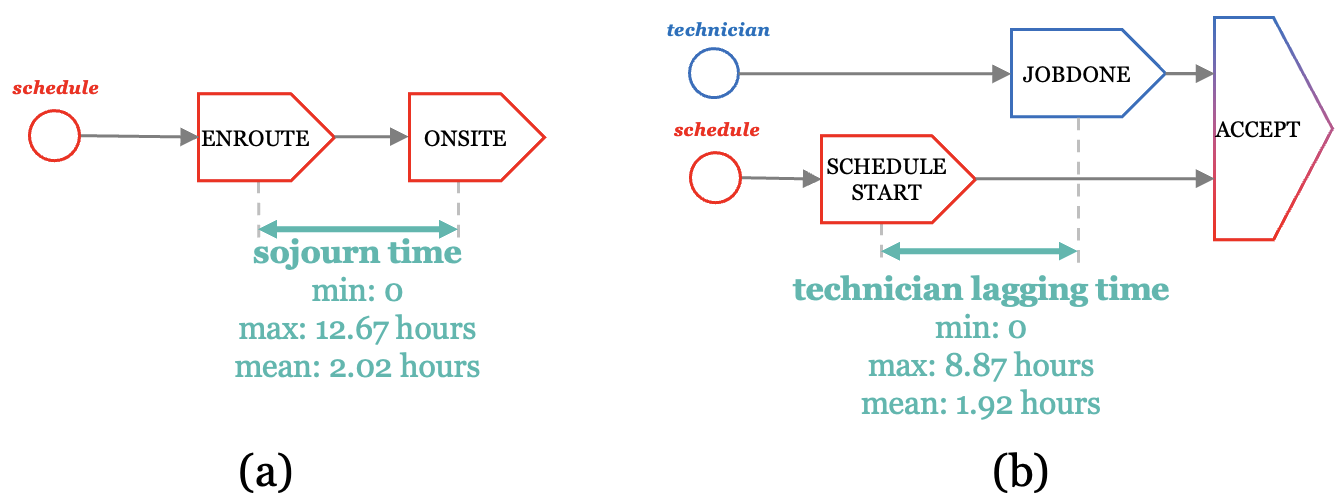}
\caption{Object-centric performance metrics: (a) transit time for a schedule and (b) a latency by technicians for accepting schedules}
\label{fig:performance}
\end{figure}

\autoref{fig:performance} shows object-centric performance metrics. 
\autoref{fig:performance}(a) measures the duration from activity ENROUTE to activity ONSITE, indicating the technician's travel time to the assigned schedules. 
With an average time of 2.02 hours --- notably long given the typical 4-hour planning of a schedule --- this metric indicates a severe bottleneck in the process.
A minimum time of zero signifies overlooked travel logging or immediate reporting upon arrival, while the upper limit of 12.67 hours hints at accidents during transit.
\autoref{fig:performance}(b) shows the lagging time of technicians for accepting schedules.
The average lagging time is 1.92 hours, which indicates the average latency of 1.92 hours caused by technicians after the schedule is already started.

\subsection{In-depth Analysis}\label{subsec:in-depth}
\begin{figure}[!ht]
\centering
\includegraphics[width=0.8\textwidth]{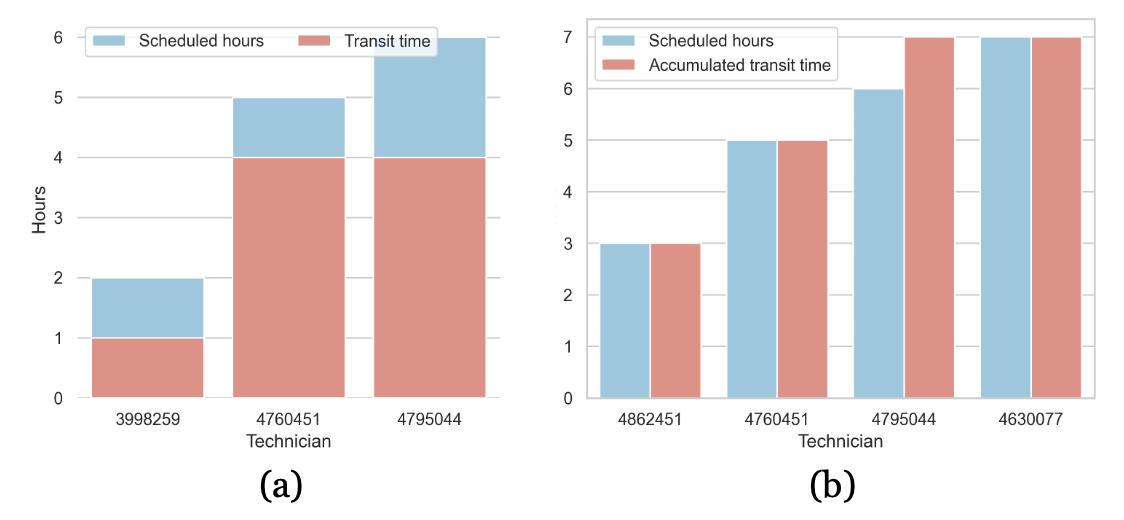}
\caption{(a) Scheduled time vs. Transit time and (b) Scheduled time vs. Accumulated transit time}
\label{fig:consequence}
\end{figure}

In this section, we delve deeper into the implications of the high transit time of technicians highlighted in the object-centric performance analysis (cf. \autoref{fig:performance}). 
As shown in \autoref{fig:consequence}(a), the transit time can account for over 50\% of a technician's scheduled time.
For example, technician 4760451's schedule was planned for 5 hours, yet 4 hours (or 80\% of the time) were dedicated solely to traveling.
Given that the original schedules allocate only a brief period for movement, such timings inevitably result in delays in schedules.

Our investigation with additional data also uncovered that technicians frequently return to the main office to gather necessary equipment and resources.
This introduces extra travel time: both from their current scheduled location to the office, and then back from the office to the next scheduled site.
As depicted in \autoref{fig:consequence}(b), the accumulated travel time can occasionally surpass the designated work schedule.

\begin{figure}[!ht]
\centering
\includegraphics[width=0.8\textwidth]{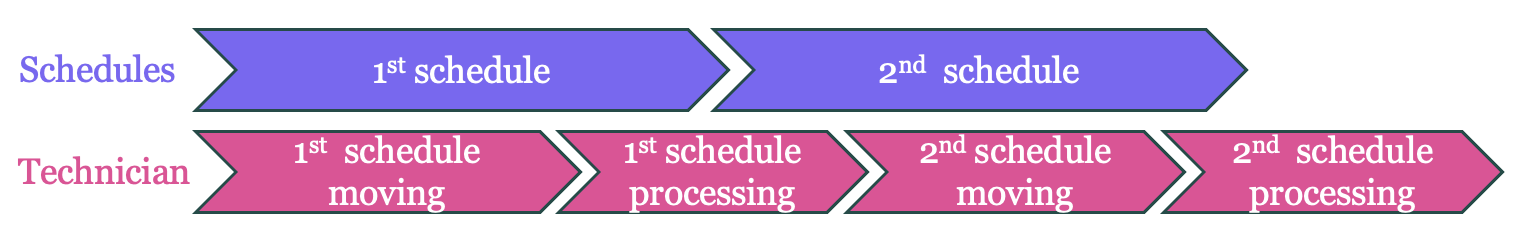}
\caption{Impact of accumulated delays on subsequent schedules: The technician started processing the second schedule almost at the end of its scheduled end.}
\label{fig:consequence2}
\end{figure}

The problem becomes even more severe when a technician has consecutive schedules. 
\autoref{fig:consequence2} showcases such a scenario.
If there is a delay in a technician's initial appointment due to prolonged travel, it not only pushes back their subsequent schedules, but the following schedules also inherit the amplified delay.

\section{Improvements}
\label{sec:improvement}
Our study led to several tangible improvements being incorporated into the organization.
We delved into issues surrounding overwork schedules (cf. \autoref{subsec:single}), uncovering root causes for technician tardiness.
As a remedy, the company initiates reminders for technicians via a mobile application, enhancing real-time process oversight.
Moreover, to prevent the consequences of high transit time (cf. \autoref{subsec:in-depth}), the company now factors in technician-customer proximity and assigns work based on anticipated processing time.

We also assess potential improvements, yet to be operationalized.
First, evaluating the disparity between overwork schedules and schedule processing times highlighted discrepancies between planned and actual technician schedules.
As a result, future schedules will incorporate historical scheduling data for more accurate planning.

Presently, once the dispatcher establishes daily schedules, they remain static, even amid day-to-day shifts.
Especially when technicians have multiple daily schedules (cf. \autoref{subsec:multi}), the introduction of a recommendation system could allow dynamic schedule updates based on real-time events.

Our research pinpointed excessive use of the ``hold'' status by technicians taking unscheduled breaks (cf. \autoref{subsec:multi}).
While this has not directly triggered any current improvement actions, it has provided valuable insights into refining the scheduling process.

\section{Conclusion}
\label{sec:conclusion}
In this case study, we explored the after-sales service process at \textit{Borusan Cat} using object-centric process mining.
The analysis has underscored the significance of intertwined processes, most notably, the interaction between schedules and technician activities.
Through examining distribution metrics, overwork schedules, and the anomalies found in daily technician activities, we have unfolded some pivotal challenges in the company – particularly the high transit times which, at times, consumed more than half of the technician's scheduled working hours. 
This led to a domino effect, where one delay translated into cascading delays across subsequent schedules, highlighting the urgent need for process optimization.

With the profound understanding derived from this case study, \textit{Borusan Cat} is now armed with the knowledge to streamline their after-sales service process.
First, the introduction of mobile reminders for technicians and geographically strategic task assignments reflect the organization's commitment to addressing the identified transit time issues.
Our insights further advocate for the use of historical data in future scheduling and the potential of recommendation algorithms for real-time adaptability.

This work has several limitations.
While our case study emphasizes two object types, i.e., schedules and technicians, the after-sales process encompasses additional objects such as work orders and order items.
A deeper analysis of the interactions among these objects could yield more comprehensive insights. 
Additionally, we have not fully harnessed the breadth of emerging techniques in object-centric process mining, such as object-centric conformance checking and variant analysis. 
Exploring these advanced methods presents a promising avenue for future research.

\section*{Acknowledgment}
The authors would like to thank the Alexander von Humboldt (AvH) Stiftung for funding this research.

\bibliographystyle{splncs04}
\bibliography{mybib}

\begin{thebibliography}{10}
\providecommand{\url}[1]{\texttt{#1}}
\providecommand{\urlprefix}{URL }
\providecommand{\doi}[1]{https://doi.org/#1}

\bibitem{Aalst16}
van~der Aalst, W.M.P.: Process Mining: Data Science in Action. Springer, 2 edn. (2016)

\bibitem{OCPM-white-paper}
van~der Aalst, W.M.P.: {Object-Centric Process Mining: The next frontier in business performance}. White paper, Celonis (March 2023)

\bibitem{math11122691}
van~der Aalst, W.M.P.: Object-centric process mining: Unraveling the fabric of real processes. Mathematics  \textbf{11}(12) (2023)

\bibitem{DBLP:journals/fuin/AalstB20}
van~der Aalst, W.M.P., Berti, A.: Discovering object-centric {Petri} nets. Fundam. Informaticae  \textbf{175}(1-4),  1--40 (2020)

\bibitem{DBLP:journals/simpa/AdamsPA22}
Adams, J.N., Park, G., van~der Aalst, W.M.P.: ocpa: {A} python library for object-centric process analysis. Softw. Impacts  \textbf{14},  100438 (2022)

\bibitem{DBLP:conf/caise/EckLLA15}
van Eck, M.L., Lu, X., Leemans, S.J.J., van~der Aalst, W.M.P.: {$PM^2$: A Process Mining Project Methodology}. In: Zdravkovic, J., Kirikova, M., Johannesson, P. (eds.) {CAiSE} 2015. Lecture Notes in Computer Science, vol.~9097, pp. 297--313. Springer (2015)

\bibitem{DBLP:books/sp/22/Fahland22}
Fahland, D.: Process mining over multiple behavioral dimensions with event knowledge graphs. In: van~der Aalst, W.M.P., Carmona, J. (eds.) Process Mining Handbook, Lecture Notes in Business Information Processing, vol.~448, pp. 274--319. Springer (2022)

\bibitem{DBLP:journals/is/GhilardiGMR22}
Ghilardi, S., Gianola, A., Montali, M., Rivkin, A.: Petri net-based object-centric processes with read-only data. Inf. Syst.  \textbf{107},  102011 (2022)

\bibitem{DBLP:conf/icpm/ParkA22}
Park, G., van~der Aalst, W.M.P.: Monitoring constraints in business processes using object-centric constraint graphs. In: Montali, M., Senderovich, A., Weidlich, M. (eds.) {ICPM} 2022. Lecture Notes in Business Information Processing, vol.~468, pp. 479--492. Springer (2022)

\bibitem{DBLP:conf/er/ParkAA22}
Park, G., Adams, J.N., van~der Aalst, W.M.P.: Opera: Object-centric performance analysis. In: Ralyt{\'{e}}, J., Chakravarthy, S., Mohania, M.K., Jeusfeld, M.A., Karlapalem, K. (eds.) {ER} 2022. Lecture Notes in Computer Science, vol. 13607, pp. 281--292. Springer (2022)

\bibitem{DBLP:conf/bpm/RebmannRA22}
Rebmann, A., Rehse, J., van~der Aa, H.: Uncovering object-centric data in classical event logs for the automated transformation from {XES} to {OCEL}. In: Ciccio, C.D., Dijkman, R.M., del{-}R{\'{\i}}o{-}Ortega, A., Rinderle{-}Ma, S. (eds.) {BPM} 2022. Lecture Notes in Computer Science, vol. 13420, pp. 379--396. Springer (2022)

\bibitem{reinkemeyer20}
Reinkemeyer, L. (ed.): Process {Mining} in {Action}: {Principles}, {Use} {Cases} and {Outlook}. Springer International Publishing (2020)

\end{thebibliography}

\end{document}